\def\year{2021}\relax
\documentclass[letterpaper]{article} 
\usepackage{aaai21}  
\usepackage{times}  
\usepackage{helvet} 
\usepackage{courier}  
\usepackage[hyphens]{url}  
\usepackage{graphicx} 
\usepackage{natbib}  
\usepackage{caption} 
\usepackage{algorithm}      
\usepackage[noend]{algorithmic}      

\usepackage{amsmath,amssymb}
\usepackage{multirow}
\usepackage{tikz}

\usepackage{amsthm}

\usepackage[switch]{lineno}  %

\title{Differentiable Inductive Logic Programming for Structured Examples}

\author {
        Hikaru Shindo\textsuperscript{\rm 1},
        Masaaki Nishino\textsuperscript{\rm 2},
        Akihiro Yamamoto\textsuperscript{\rm 1} \\
}
\affiliations {
    \textsuperscript{\rm 1} Kyoto University, Japan \\
    \textsuperscript{\rm 2} NTT Communication Science Laboratories, Japan \\
    shindo@iip.ist.i.kyoto-u.ac.jp, masaaki.nishino.uh@hco.ntt.co.jp, akihiro@i.kyoto-u.ac.jp
}

\begin{document}
\maketitle

\begin{abstract}
The differentiable implementation of logic yields a seamless combination of symbolic reasoning and deep neural networks.
Recent research, which has developed a differentiable framework to learn logic programs from examples, can even acquire reasonable solutions from noisy datasets.
However, this framework severely limits expressions for solutions, e.g., no function symbols are allowed, and the shapes of clauses are fixed.
As a result, the framework cannot deal with structured examples.
Therefore we propose a new framework to learn logic programs from noisy and structured examples, including the following contributions.
First, we propose an adaptive clause search method by looking through structured space, which is defined by the generality of the clauses, to yield an efficient search space for differentiable solvers.
Second, we propose for ground atoms an enumeration algorithm, which determines a necessary and sufficient set of ground atoms to perform differentiable inference functions.
Finally, we propose a new method to compose logic programs softly, enabling the system to deal with complex programs consisting of several clauses.
Our experiments show that our new framework can learn logic programs from noisy and structured examples, such as sequences or trees.
Our framework can be scaled to deal with complex programs that consist of several clauses with function symbols.
\end{abstract}

\section{Introduction}
\noindent 
Integrating symbolic reasoning and numerical computation is increasingly becoming a vital factor in artificial intelligence and its applications~\cite{DeRaedt20}.
Due to the success of deep neural networks (DNNs), one of the main integrated techniques is to combine DNNs with logical reasoning, which is called neuro-symbolic computation~\cite{garcez19}.
The main goal is to establish a unified framework that can make flexible approximations using DNNs and perform tractable and multi-hop reasoning using first-order logic.

Although many approaches have been developed for the integration of logic and DNNs ~\cite{rocktaschel17,Yang17,Sourek18,Manhaeve19,Si19,Cohen20,Riegel20,Marra20},
most existing approaches involve the learning of continuous parameters, not discrete structures.
Structure learning~\cite{Kok05}, in which logical expressions are obtained explicitly, presents a challenge to neuro-symbolic approaches~\cite{DeRaedt20}.

Evans and Grefenstette proposed~\cite{Evans18} Differentiable Inductive Logic Programming ($\partial$ILP), which is a framework for learning logic programs from given examples in a differentiable manner.
Inductive Logic Programming (ILP)~\cite{Muggleton91} is a sound formalization for finding theories from given examples using first-order logic as its language~\cite{Nienhuys97}.
The $\partial$ILP framework formulates ILP problems as numerical optimization problems that can be solved by gradient descent.
Its differentiability establishes a seamless combination of ILP and neural networks to deal with subsymbolic and noisy data.

However, previous work has put severe limitations on expressions for solutions.
For instance, no function symbols are allowed, the arity of predicates must be less than $2$, the number of atoms in the clause body must not exceed $2$, and every program must be comprised of pairs of rules for each predicate.
Thus it is unsuitable for complex structured data, such as sequences or trees, or complex programs that are comprised of several clauses for a predicate.
One main characteristic of first-order logic is the expressibility and learnability for structured data with function symbols~\cite{Lloyd03,Dantsin01,Fredouille07}.
We face $three$ main challenges to deal with complex programs and structured data:
(i) the number of clauses to be considered increases,
(ii) an infinite number of ground atoms can be generated with function symbols, 
and (iii) the memory and computation costs increase quadratically with respect to the size of the search space.
We address these issues by 
proposing a new differentiable approach to learning logic programs by combining adaptive symbolic search methods and continuous optimization methods and make the following contributions for each problem:
\\
    {\bf Clause Search with Refinement} We propose an efficient clause search method for a differentiable ILP solver.
    We generate clauses by beam searching and leveraging the generality of clauses and the given examples.
    We start from general (strong) clauses and incrementally specify (weaken) the clauses. We only take clauses that contribute to accurate classification results into the search space.
    Our approach yields an efficient search space that includes only promising clauses for the differentiable ILP solver.
    \\
    {\bf Adaptive Fact Enumeration} We present a fact enumeration algorithm to implement the differentiable inference function.  
    The number of ground atoms defines the size of the tensors used in the differentiable step, thus the set of required ground atoms must be determined.
    We enumerate the ground atoms using the given examples and the generated clauses by backward-chaining. 
    Our approach yields a small set of ground atoms, and this small set is a key factor to achieve differentiable learning from structured objects. 
    \\
    {\bf Soft Program Composition} We propose a practical algorithm to learn complex logic programs in a differentiable manner. 
    In past studies, the weights were assigned to each pair of clauses because some information is lost if weights are assigned to each clause, and thus the number of parameters increased quadratically.
    In our approach, we compose a differentiable inference function by assigning multiple distinct weights to each clause and introducing a function to compute logical {\em or} softly.
    Our approach efficiently estimates logic programs in terms of memory and computation costs.

\vspace{0.5em}
\noindent {\bf Notation}
We use bold lowercase letters $\bf{v}, \bf{w}, \ldots$ for vectors and the functions that return vectors.
We use bold capital letters $\bf{X}, \ldots$ for tensors.
We use calibrate letters $\mathcal{C}, \mathcal{A}, \ldots$ for sets and ordered sets.

\section{Related Work}
A  pioneering study of inductive inference was done in the early 70s~\cite{Plotkin71}.
The Model Inference System (MIS)~\cite{Shapiro83} has been implemented as an efficient search algorithm for logic programs using the generality of expressions.
Inductive Logic Programming~\cite{Muggleton91} has emerged at the intersection of machine learning and logic programming.
The Elementary Formal System (EFS)~\cite{arikawa1} is a well-established system for strings based on first-order logic.

Dealing with uncertainty in ILP has been a major obstacle.
Probabilistic Inductive Logic Programming~\cite{DeRaedt04} combines probability theory with ILP.
It is also known as Statistical Relational Learning~\cite{DeRaedt16}.
Another approach to cope with uncertainty is to combine neural methods with differentiable implementations of logic~\cite{rocktaschel17,Yang17,Evans18,Sourek18,Manhaeve19,Si19,Riegel20,Cohen20,Marra20}.
Both of these approaches blazed a trail for the integration of logic, probability, and neural methods.
However, almost all of these approaches are domain-specific~\cite{DeRaedt20}, i.e., the expressions are severely limited.
A critical gap exists between these past approaches and logic-based systems for structured data, such as MIS and EFS.
Our work fills the gap by incorporating symbolic methods with differentiable approaches.

A propositional approach for ILP is one established approach, which was developed to integrate ILP and SAT solvers or Binary Decision Diagrams~\cite{Chikara15,Shindo18}.
The $\partial$ILP system performed differentiable learning by incorporating continuous relaxation into these approaches. We also follow this approach.

Beam searching with clause refinement was developed for structure learning for probabilistic logic programs~\cite{Bellodi15,Nguembang19}.
We use this approach because it requires fewer declarative biases than approaches based only on templates.


\section{Inductive Logic Programming Concepts}
\paragraph{Basic Concepts}
{\it Language} $\mathcal{L}$ is a tuple $(\mathcal{P}, \mathcal{F}, \mathcal{A}, \mathcal{V})$,
where $\mathcal{P}$ is a set of predicates, $\mathcal{F}$ is a set of function symbols, $\mathcal{A}$ is a set of constants, and $\mathcal{V}$ is a set of variables.
We denote $n$-ary predicate $p$ by $p/n$ and $n$-ary function symbol $f$ by $f/n$.
A {\it term} is a constant, a variable, or an expression $f(t_1, \ldots, t_n)$ where $f$ is a $n$-ary function symbol and $t_1, \ldots, t_n$ are terms.
A function symbol yields a structured expression.
An {\it atom} is a formula $p(t_1, \ldots, t_n)$, where $p$ is an $n$-ary predicate symbol and $t_1, \ldots, t_n$ are terms.
A {\it ground atom} or simply a {\it fact} is an atom with no variables.
A {\it literal} is an atom or its negation.
A {\it positive literal} is just an atom. 
A {\it negative literal} is the negation of an atom.
A {\it clause} is a finite disjunction ($\lor$) of literals. 
A {\it definite clause} is a clause with exactly one positive literal.
If  $A, B_1, \ldots, B_n$ are atoms, then $A \lor \lnot B_1 \lor \ldots \lor \lnot B_n$ is a definite clause.
We write definite clauses in the form of $A \leftarrow B_1 \land \ldots \land B_n$.
Atom $A$ is called the {\it head}, and set of negative atoms $\{B_1, \ldots, B_n\}$ is called the {\it body}.
We denote special constant $\mathit{true}$ as $\top$ and  $\mathit{false}$ as $\bot$.
We denote a set of variables in clause $C$ as $V(C)$.
$DV_n(C)$ is a set of all $n$-combinations of distinct variables in clause $C$, i.e., $DV_n(C) = \{ (x_1, \ldots, x_n) | x_i \in V(C) \land x_i \neq x_j (i \neq j) \}$.
Substitution $\theta = \{x_1 = t_1, ..., x_n = t_n\}$ is an assignment of term $t_i$ to variable $x_i$. An application of substitution $\theta$ to atom $A$ is written as $A \theta$.
A {\it unifier} for the set of expressions $\{ A_1, \ldots, A_n \}$ is a substitution $\theta$ such that $A_1\theta = A_2\theta = \ldots =A_n\theta$, written as $\theta = \sigma(\{A_1, \ldots, A_n\})$, where $\sigma$ is a {\it unification function.}
A unification function returns the (most general) unifier for the expressions if they are unifiable.
Decision function $\bar{\sigma}(\{A_1, \ldots, A_n \})$ returns a Boolean value whether or not $A_1, \ldots, A_n$ are unifiable.

\subsection{Inductive Logic Programming}
ILP problem $\mathcal{Q}$ is tuple $(\mathcal{E}^+, \mathcal{E}^-, \mathcal{B}, \mathcal{L})$, where
$\mathcal{E}^+$ is a set of positive examples, $\mathcal{E}^-$ is a set of negative examples, $\mathcal{B}$ is background knowledge, 
and $\mathcal{L}$ is a language. We assume that the examples and the background knowledge are ground atoms.
The solution to an ILP problem is a set of definite clauses $\mathcal{H} \subseteq \mathcal{L}$
that satisfies the following conditions:
\begin{itemize}
    \item $\forall A \in \mathcal{E}^+ ~ \mathcal{H} \cup \mathcal{B}  \models A$.
    \item $\forall A \in \mathcal{E}^-  ~\mathcal{H} \cup \mathcal{B} \not \models A.$
\end{itemize}
Typically the search algorithm starts from general clauses. If the current clauses are too general (strong), i.e., they entail too many negative examples, then the solver incrementally specifies (weakens) them.
This weakening operation is called a {\it refinement}, which is one of the essential tools for ILP.

\noindent {\bf Refinement Operator}
The refinement operator defines between clauses the complexity that varies from {\em general} to {\em specific}. 
The refinement operator takes a clause and returns weakened clauses.
Generally, there are four types of refinement operators: (i) application of function symbols, (ii) substitution of constants, (iii) replacement of variables, and (iv) addition of atoms.
For clause $C = A \leftarrow B_1, \ldots, B_m$, each refinement operation for language $\mathcal{L} = (\mathcal{P}, \mathcal{F}, \mathcal{A}, \mathcal{V})$ is as follows:
\begin{itemize}
    \item For $z \in V(C)$, $f \in \mathcal{F}$, and $x_1, \ldots, x_n \in  \mathcal{V} \backslash V(C)$, let $C\{ z = f(x_1, \ldots, x_n) \} \in \rho_\mathcal{L}^\mathit{fun}(C)$, where $x_1, \ldots, x_n$ are pairwise different.
    \item For $z \in V(C)$ and $a \in \mathcal{A}$, let $C\{z=a \} \in \rho_\mathcal{L}^\mathit{sub}(C) $.
    \item For $z,y \in V(C) ~(z \neq y)$, let $C\{z=y\} \in \rho_\mathcal{L}^\mathit{rep}(C)$.
    \item For $n$-ary predicate $p \in \mathcal{P}$ and $(x_1, \ldots, x_n) \in DV_n(C)$, let $ A \leftarrow B_1, \ldots, B_m, p(x_1, \ldots, x_n) \in \rho_\mathcal{L}^\mathit{add}(C)$.
\end{itemize}

\noindent The refinement operator for language $\mathcal{L}$ is defined:
\begin{align}
    \rho_\mathcal{L}(C) = \rho_\mathcal{L}^\mathit{func}(C) \cup \rho_\mathcal{L}^\mathit{subs}(C) \cup \rho_\mathcal{L}^\mathit{rep}(C) \cup \rho_\mathcal{L}^\mathit{add}(C).
\end{align}
{\bf Example 1}
Let $\mathcal{L} = (\mathcal{P}, \mathcal{F}, \mathcal{A}, \mathcal{V})$, where $\mathcal{P} = \{p/2, q/2 \}$, $\mathcal{F} = \{f/1\}$, $\mathcal{A} = \{ a, b \}$ and $\mathcal{V}=\{ x,y,z\}$.
Let $\mathcal{E}^+ = \{p(a,a),$ $p(b,b) \}$, $\mathcal{E}^- = \{p(a,b), p(b,a) \}$, $\mathcal{B} = \{ \}$.
One of the solutions is $\mathcal{H} = \{ p(x,x) \}$.
\\
{\bf Example 2}
Let $\mathcal{L}$ be the language specified in Example 1.
The following is the result of the refinement:
$\rho_\mathcal{L}(p(x,y)) = \{ p(a,y)$, $p(x,a)$, $p(b,y)$, $p(x,b)$ $p(x,x)$,  $p(f(z), y)$, $p(x, f(z))$,  $p(x,y) \leftarrow q(x,y)$  $ \}$.

\subsection{Differentiable Inductive Logic Programming}
In the $\partial$ILP framework~\cite{Evans18}, an ILP problem is formulated as an optimization problem that has the following general form:
\begin{align}
    \min_\mathcal{W} L(\mathcal{Q}, \mathcal{C}, \mathcal{W}),
\end{align}
where $\mathcal{Q}$ is an ILP problem, $\mathcal{C}$ is a set of clauses specified by templates, $\mathcal{W}$ is a set of weights for clauses, and $L$ is a loss function that returns a penalty when training constraints are violated.
We briefly summarize the steps of the process as follows:
\\
{\bf Step 1}~ 
Set of ground atoms $\mathcal{G}$ is specified by given language $\mathcal{L} \in \mathcal{Q}$. 
\\
{\bf Step 2}~ Tensor ${\bf X}$ is built from given set of clauses $\mathcal{C}$ and fixed set of ground atoms $\mathcal{G}$. It holds the relationships between clauses $\mathcal{C}$ and ground atoms $\mathcal{G}$. Its dimension is proportional to $|\mathcal{C}|$ and $|\mathcal{G}|$.
\\
{\bf Step 3}~ Given background knowledge $\mathcal{B} \in \mathcal{Q}$ is compiled into vector ${\bf v}_0 \in \mathbb{R}^{|\mathcal{G}|}$.
Each dimension corresponds to each ground atom $G_j \in \mathcal{G}$, and ${\bf v}_0[j]$ represents the valuation of $G_j$.
\\
{\bf Step 4}~ A computational graph is constructed from ${\bf X}$ and $\mathcal{W}$.
 The weights define probability distributions over clauses $\mathcal{C}$.
 A probabilistic forward-chaining inference is performed by the forwarding algorithm on the computational graph with input ${\bf v}_0$. 
\\
{\bf Step 5}~ 
The loss is minimized with respect to weights $\mathcal{W}$ by gradient descent techniques.
After minimization, a human-readable program is extracted by discretizing the weights.
\begin{figure}
    \centering
    \includegraphics[width=\linewidth]{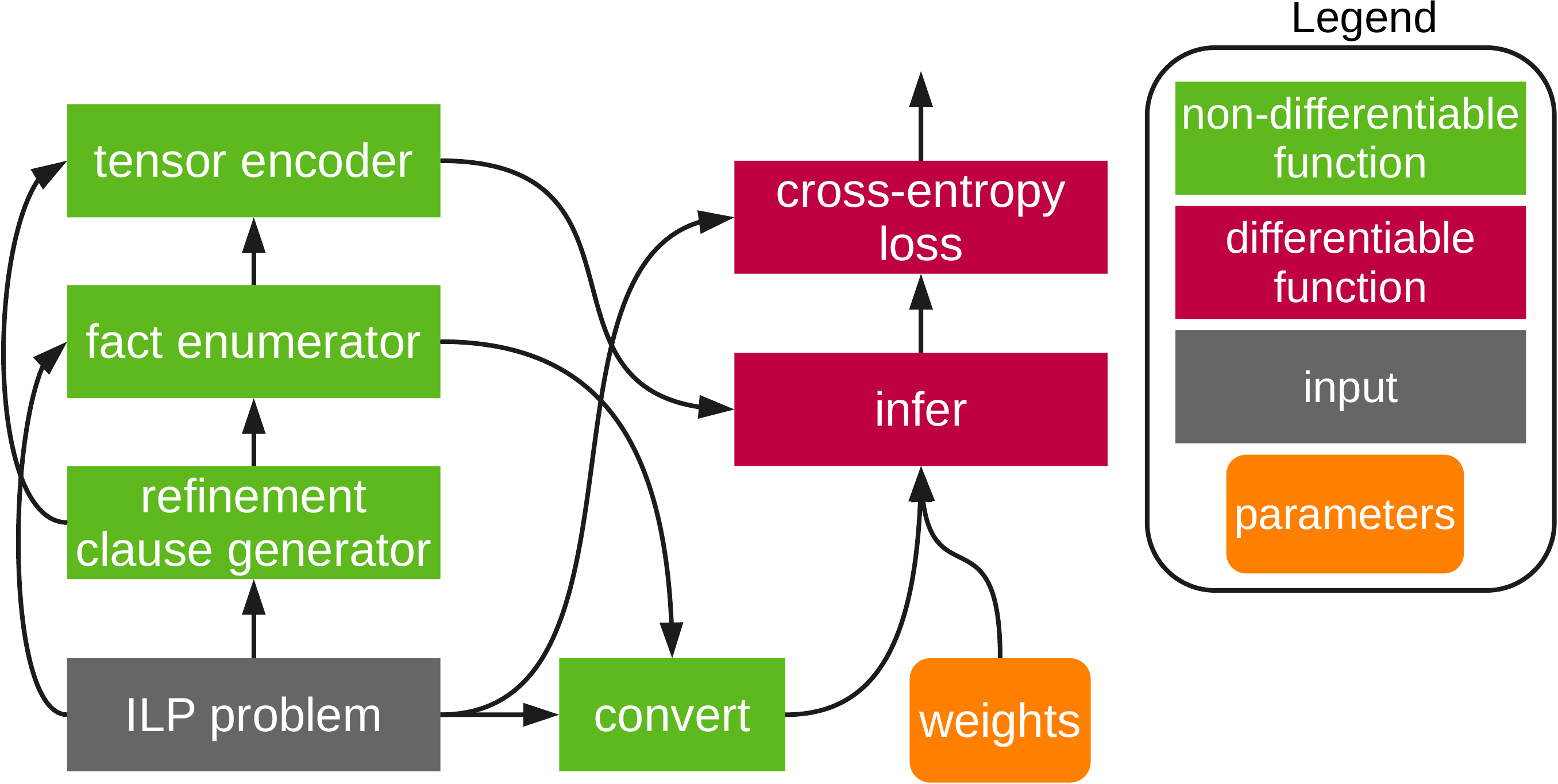}
    \caption{Overview of our model}
    \label{fig:flow}
\end{figure}

\section{Method}
Although we begin by following the $\partial$ILP approach, we introduce several new algorithms to deal with structured examples and complex programs with function symbols.
An overview of our approach is illustrated in Fig. \ref{fig:flow}.
First, we generate clauses by beam searching with refinement to specify an efficient search space.
Second, we enumerate ground atoms by backward-chaining using the set of generated clauses.
This enumeration results in efficient inference computation because the number of ground atoms determines the dimensions of tensors for the differentiable steps.
Third, we propose a new approach to softly compose complex logic programs.
We assign several weights for each clause to define several probability distributions over the clauses and efficiently estimate logic programs that consist of several clauses.

\subsection{Clause Search with Refinement}
We incrementally generate candidates of clauses by refinement and beam searching.
Promising clauses for an ILP problem are those that entail many positive examples but few negative examples.
Algorithm 1 is our generation algorithm.
The inputs are initial clauses $\mathcal{C}_0$, ILP problem $\mathcal{Q}$, the size of the beam in search $N_\mathit{beam}$, and the number of steps of beam searching $T_\mathit{beam}$.
We start from the initial clauses and iteratively weaken the top-$N_\mathit{beam}$ clauses based on how many positive examples can be entailed by clause combining with background knowledge.
The following is the evaluation function for clause $R$:
\begin{align}
    \mathit{eval}(R, \mathcal{Q}) = |\{E ~|~ E \in \mathcal{E}^+ \land \mathcal{B} \cup \{R\} \models E \} |,
\end{align}
where $\mathcal{E}^+$ is a set of positive examples.

\begin{figure}[t]
\begin{center}
\begin{tikzpicture}[nodes={}, ->, scale=0.80, transform shape]
\node{$^*p(x,y)$}
[sibling distance=2.2cm]
    child { node {$p(a,y)$} }
    child { node {$^*p(x,x)$} 
            [sibling distance=1.2cm]
            child {node {$\vdots$}}
            child {node {$\vdots$}}
            child {node {$\vdots$}}}
    child { node {$p(f(x), y)$}}
    child { node {$^*p(x,y) \leftarrow q(x,y)$} 
            [sibling distance=1.3cm]
            child {node {$\vdots$}}
            child {node {$\vdots$}}
            child {node {$\vdots$}}};
\end{tikzpicture}
\caption{Beam searching for clauses}
\label{fig:beam_search}
\end{center}
\end{figure}
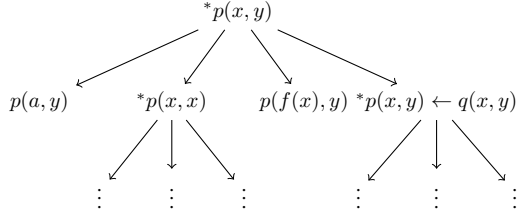

The key difference from $\partial$ILP is that we leverage the given examples to specify the search space for the differentiable solver. 
In $\partial$ILP, since the clauses are generated only by templates many meaningless clauses tend to be generated.
\\
\noindent {\bf Example 3}
Let $\mathcal{E}^+ = \{p(a,a),$ $p(b,b),$ $p(b,c),$ $p(c,b) \}$, $\mathcal{B} = \{q(b,c), q(c,b) \}$, $\mathcal{C}_0 = \{ p(x,y)\}$, $T_\mathit{beam} = 2$, and $N_\mathit{beam}=2$.
Fig. \ref{fig:beam_search} illustrates an example of beam searching for this problem.
In the 2nd layer, we show examples of generated clauses by refining the initial clause.
Each new clause is evaluated and selected to be refined. In this case, clause $p(x,x)$ and $p(x,y)\leftarrow q(x,y)$ is refined in the next step 
because it entails more positive examples with background knowledge $\mathcal{B}$ than other clauses.
Refined clauses are added to set $\mathcal{C}$.
By contrast, since clause $p(f(x),y)$ does not entail any positive examples with background knowledge $\mathcal{B}$, it is discarded.
Finally, we get set of clauses $\mathcal{C} = \{ p(x,y), p(x,x), p(x,y)\leftarrow q(x,y)\}$.

\begin{algorithm}[t]
\caption{Clause generation by beam searching}
\begin{algorithmic}[1]
\small
\REQUIRE  $\mathcal{C}_0, \mathcal{Q}, N_\mathit{beam}, T_\mathit{beam}$
\STATE $\mathcal{C}_\mathit{to\_open} \leftarrow \mathcal{C}_0$
\STATE $\mathcal{C} \leftarrow \emptyset$
\STATE $t = 0$
    \WHILE {$t < T_\mathit{beam}$}
    \STATE $\mathcal{C}_\mathit{beam} \leftarrow \emptyset$
        \FOR{ $C_i \in \mathcal{C}_\mathit{to\_open}$}
            \STATE $\mathcal{C} = \mathcal{C} \cup \{C_i\}$
            \FOR{$R \in \rho_\mathcal{L}(C_i)$}
                \STATE $\mathit{score} = \mathit{eval}(R,\mathcal{Q})$ //Evaluate each clause
                \STATE $\mathcal{C}_\mathit{beam} = \mathit{insert}(\mathcal{C}_\mathit{beam}, R, \mathit{score})$ //Insert refined clause in order of scores possibly discarding it
            \ENDFOR
        \ENDFOR
    \STATE $\mathcal{C}_\mathit{to\_open} = \mathcal{C}_\mathit{beam}$ //top-$N_\mathit{beam}$ clauses are refined in the next loop
    \STATE $t = t + 1$
    \ENDWHILE
    \RETURN $\mathcal{C}$
\end{algorithmic}
\end{algorithm}

\subsection{Adaptive Fact Enumeration}
We enumerate ground atoms using the given clauses and examples.
Algorithm 2 is our enumeration algorithm.
The inputs are ILP problem $\mathcal{Q}=$ $(\mathcal{E}^+, \mathcal{E}^-, \mathcal{B}, \mathcal{L})$, set of clauses $\mathcal{C}$, and time-step parameter $T$ that determines the number of forward-chaining steps in the differentiable inference.
We start from the given examples, the background knowledge, and special symbols that represent {\em true} and {\em false} respectively.
We unify the head of each clause and each ground atom. 
If they are unifiable, then we compute the ground atoms on the body by applying the unifier.
Here we assume that the body has fewer variables than the head. 

The key difference from $\partial$ILP is that we utilize the given ILP problem to specify the set of ground atoms.
In $\partial$ILP, the solver considers all the visible ground atoms, which is known as the {\em Herbrand Base}. 
However, since an infinite number of ground atoms are yielded by function symbols, it is unsuitable for the case with function symbols.

\begin{algorithm}[t]
\caption{Enumeration of ground atoms}
\begin{algorithmic}[1]
\REQUIRE $\mathcal{Q}$, $\mathcal{C}$, $T$
\small
\STATE $\mathcal{G} \leftarrow \{\bot, \top \} \cup \mathcal{E}^+ \cup \mathcal{E}^- \cup \mathcal{B}$
  \FOR {$i = 0$ to $T-1$}
    \STATE $\mathcal{S} \leftarrow \emptyset$
    \FOR {$A \leftarrow B_1, \ldots, B_n $ in  $\mathcal{C}$}
      \FOR {$G \in \mathcal{G}$}
       \IF {$\bar{\sigma}(A,G)$}
       \STATE  $\theta \leftarrow \sigma(A, G)$\\
        $\mathcal{S} \leftarrow \mathcal{S} \cup \{ B_1\theta, \ldots, B_n\theta \}$
       \ENDIF
      \ENDFOR
    \ENDFOR
    \STATE $\mathcal{G} \leftarrow \mathcal{G} \cup \mathcal{S}$
  \ENDFOR
 \RETURN $\mathcal{G}$ 
\end{algorithmic}
\end{algorithm}

\noindent {\bf Example 4}
    Let $\mathcal{E}^+ = \{e(s^6(0)) \},$ $\mathcal{E}^- = \{e(s(0)) \}$, $ \mathcal{B} = \{ e(0)\}$,
    $\mathcal{C} = $ $\{ e(s^2(x))$ $\leftarrow e(x) \}$, and $T = 2$.
    First $\mathcal{G}$ is initialized as $\mathcal{G} =$ $\{ \bot, \top,$ $e(s^6(0)), e(s(0)), e(0) \}$.
    Atom $e(s^6(0))$ and clause head $e(s^2(x))$ are unifiable with $\theta =$ $\{ x=s^4(0)\}$. 
    Then body $e(x)\theta =$ $e(s^4(0))$, and this ground atom is added to $\mathcal{G}$.
    In the next step, atom $e(s^4(0))$ and clause head $e(s^2(x))$ are unifiable with $\theta=\{ x = s^2(0)\}$.
    Hence body $e(x) \theta = e(s^2(0))$ is added to $\mathcal{G}$.
    Finally, the enumeration algorithm returns $\mathcal{G} = \{\bot, \top, e(0), e(s(0)), e(s^2(0)), e(s^4(0)), e(s^6(0)) \}$. 
   Note that ground atoms $e(s^3(0))$ and $e(s^5(0))$ are not required in this case.

\subsection{Soft Program Composition}
\paragraph{Tensor Encoding}
We build a tensor that holds the relationships between clauses $\mathcal{C}$ and ground atoms $\mathcal{G}$.
We assume that $\mathcal{C}$ and $\mathcal{G}$ are an ordered set, i.e., where every element has its own index.
Let $b$ be the maximum body length in $\mathcal{C}$.
Index tensor ${\bf X} \in \mathbb{N}^{|\mathcal{C}| \times |\mathcal{G}| \times b}$ contains the indexes of the ground atoms to compute forward inferences.
Intuitively, ${\bf X}[i,j] \in \mathbb{N}^b$ contains a set of the indexes of the subgoals to entail the $j$-th fact using the $i$-th clause.
For clause $C_i = A \leftarrow B_1, \ldots, B_n \in \mathcal{C}$ and set of ground atoms $\mathcal{G}$, we compute tensor ${\bf X}$:
\begin{align}
    {\bf X}[i,j,k] = \begin{cases} I_\mathcal{G}(B_k \theta) ~\mbox{if}~ \bar{\sigma}(\{A, G_j \}) \land k \leq n\\
    I_\mathcal{G}(\top) ~\mbox{if}~ \bar{\sigma}(\{A, G_j \}) \land k > n\\
    I_\mathcal{G}(\bot) ~\mbox{if}~ \lnot \bar{\sigma}(\{A, G_j \})
  \end{cases},
  \label{eq:build_tensor}
\end{align}
where
$0 \leq j \leq |\mathcal{G}|-1$, $0 \leq k \leq b-1$, $\theta = \sigma(\{A, G_j\})$, and $I_\mathcal{G}(x)$ returns the index of $x$ in $\mathcal{G}$.
If clause head $A$ and ground atom $G_j$ are unifiable, then we put the index of subgoal $B_k \theta$ into the tensor (line 1 in Eq. \ref{eq:build_tensor}).
If the clause has fewer body atoms than the longest clause in $\mathcal{C}$, we fill the gap with the index of $\top$ (line 2 in Eq. \ref{eq:build_tensor}).
If clause head $A$ and ground atom $G_j$ are not unifiable, then we place the index of $\bot$ (line 3 in Eq. \ref{eq:build_tensor}).

\noindent {\bf Example 5}
Let $C_0 = e(x), C_1 = e(s^2(x)) \leftarrow e(x)$ 
and $\mathcal{G} =$ $\{\bot, \top, $ $e(0), e(s(0)), $ $e(s^2(0)), e(s^4(0)) \}$. 
Then the following table shows tensor ${\bf X}$:
\begin{table}[H]
\footnotesize
    \centering
    \begin{tabular}{c|cccccc}
    \hline
    $j$ &0 & 1 & 2 & 3& 4&5\\\
    $\mathcal{G}$ & $\bot$ & $\top$ & $e(0)$ & $e(s(0))$ & $e(s^2(0))$ &$e(s^4(0))$  \\\hline\hline
    ${\bf X}[0,j]$ & $[0]$ & $[1]$ & $[1]$ & $[1]$ & $[1]$ & $[1]$  \\
    ${\bf X}[1,j]$ & $[0]$ & $[1]$ & $[0]$ & $[0]$ & $[2]$ & $[4]$  \\
    \end{tabular}
\end{table}
\noindent For example, ${\bf X}[1,4] = [2]$ because clause $C_1$ entails $e(s^2(0))$ with substitution $\theta = \{ x=0\}$.
Then subgoal $e(x) \theta = e(0)$, which has index $2$. 
Clause $C_0$ does not have a body atom, and so the body is filled by $\top$, which has index $1$.

\paragraph{Valuation}
Valuation vector  ${\bf v}_t \in \mathbb{R}^{|\mathcal{G}|}$ maps each ground atom into a continuous value at each time step $t$.
The background knowledge is compiled into ${\bf v}_0$:
\begin{align}
    {\bf v}_0[j] = {\bf f}_\mathit{convert}(\mathcal{B})[j] =  \begin{cases}
    1 ~ (G_j \in \mathcal{B} \lor G_j = \top)\\
    0 ~ (\mbox{otherwise})
    \end{cases}.
\end{align}
The differentiable inference function is performed based on valuation vectors.
To compute the $T$-step forward-chaining inference, we compute the sequence of valuation vectors ${\bf v}_0, \ldots, {\bf v}_T$ in the differentiable inference process.

\paragraph{Clause Weights}
We assign weights to softly compose the logic programs as follows:
(i) We fix the target programs' size as $m$, i.e., where we try to find a logic program with $m$ clauses.
(ii)  We introduce $|\mathcal{C}|$-dim weights $\mathcal{W} = \{ {\bf w}_1, \ldots, {\bf w}_m \}$.
(iii) We take the softmax of each weight vector ${\bf w}_l \in \mathcal{W}$ and softly choose $m$ clauses to compose the logic program.
As a probabilistic interpretation, we define a probability distribution $p(x_i^l)$, where $x_i^l$ is a probabilistic variable representing clause $C_i$ is the $l$-th component of the target program. 

In $\partial$ILP, the weights are assigned to each pair of clauses by assuming all programs are composed of pairs of clauses for each predicate.
In our method, we assign several weights to each clause and softly choose each clause.
Our approach enables the solver to deal with complex programs that consist of several clauses with identical predicates.

\paragraph{Differentiable Inference}
We compose a differentiable function, called an {\em infer function}, that performs forward-chaining inference.
The inference result is obtained:
\begin{align}
    {\bf v}_T = {\bf f}_\mathit{infer}({\bf X}, {\bf v}_0, \mathcal{W}, T),
\end{align}
where 
${\bf f}_\mathit{infer}$ is the infer function, ${\bf X}$ is the index tensor, ${\bf v}_0$ is the initial valuation vector, 
$\mathcal{W}$ is the set of weight vectors, and $T$ is the time step.

The infer function is computed as follows.
First, each clause $C_i \in \mathcal{C}$ is compiled into a function ${\bf c}_i:\mathbb{R}^{|\mathcal{G}|} \rightarrow \mathbb{R}^{|\mathcal{G}|}$:
\begin{align}
   {\bf c}_{i}({\bf v}_t)[j] = \prod_k {\bf gather}({\bf v}_t, {\bf X}[i])[j,k],
\end{align}
where function ${\bf gather}: \mathbb{R}^{|\mathcal{G}|} \times \mathbb{N}^{|\mathcal{G}|\times b} \rightarrow \mathbb{R}^{|\mathcal{G}| \times b}$ is:
\begin{align}
    {\bf gather}({\bf a}, {\bf B})[j,k] = {\bf a}[{\bf B}[j,k]]. 
\end{align}
The ${\bf gather}$ function replaces the indexes of the ground atoms by the current valuation values.
To take logical {\em and} across the subgoals in the body, we take the product across dimension $1$.

Next we take the weighted sum of the clause function using ${\bf w}_l \in \mathcal{W}$: 
\begin{align}
     {\bf h}_l({\bf v}_t) = \sum_i {\bf softmax}({\bf w}_l)[i] \cdot {\bf c}_{i}({\bf v}_t),
\end{align}
where ${\bf softmax}({\bf x})[i] = \frac{\exp({\bf x}[i])}{\sum_{i^\prime} \exp({\bf x}[i^\prime])}$.
Note that ${\bf softmax}({\bf w}_l)[i]$ is interpreted as a probability that $C_i \in \mathcal{C}$ is the $l$-th component of the target program.

Then we compute the forward-chaining inference using clauses $\mathcal{C}$ and weights $\mathcal{W}$:
\begin{align}
    {\bf r}({\bf v}_t) = {\bf softor}^\gamma \left({\bf h}_{1}({\bf v}_t), \ldots, {\bf h}_{m}({\bf v}_t)  \right),
\end{align}
where ${\bf softor}^\gamma$ is a smooth logical {\em or}  function on the valuation vectors:
\begin{align}
    {\bf softor}^\gamma({\bf x}_1, \ldots, {\bf x}_m)[j] = \gamma \log \sum_l e^{{\bf x}_l[j] / \gamma},
\end{align}
where $\gamma > 0$ is a smooth parameter.
Taking logical {\em or} softly for the valuation vectors corresponds to the fact that a logic program is generally represented as a conjunction of clauses.

Finally, we perform $T$-step inference by iteratively amalgamating the results:
\begin{align}
    {\bf v}_{t+1} &= {\bf softor}^\gamma\left( {\bf v}_t, {\bf r}({\bf v}_t)\right).
    \label{eq:update}
\end{align}
Infer function ${\bf f}_\mathit{infer}({\bf X}, {\bf v}_0, \mathcal{W}, T)$ returns ${\bf v}_T$.

\subsection{Learn Target Program}
Let $\mathcal{Q} = (\mathcal{E}^+, \mathcal{E}^-, \mathcal{B}, \mathcal{L})$.
We generate pairs of atoms and labels as:
\begin{align}
  \mathcal{Y} = \{(E, 1) ~|~ E \in \mathcal{E}^+ \} \cup \{(E, 0) ~|~ E \in \mathcal{E}^- \}.
\end{align}
Each pair $(E, y)$ represents whether atom $E$ is positive or negative.
We compute the conditional probability of label $y$ of atom $E$: 
\begin{align}
    p(y~|~ E, \mathcal{Q}, \mathcal{C}, \mathcal{W}, T) = {\bf f}_\mathit{infer}({\bf X}, {\bf v}_0, \mathcal{W}, T)[I_\mathcal{G}(E)],
\end{align}
where
$\mathcal{C} =$ $f_\mathit{beam\_search}(\mathcal{C}_0, \mathcal{Q},$ $ N_\mathit{beam}, T_\mathit{beam})$,
$\mathcal{G} =$  $f_\mathit{enumerate}(\mathcal{C}, \mathcal{Q}, T)$,
${\bf v}_0 =$ ${\bf f}_\mathit{convert}(\mathcal{B})$,
 ${\bf X}$ is the index tensor,
 and $I_\mathcal{G}(x)$ returns the index of $x$ in $\mathcal{G}$.
Here 
$f_\mathit{beam\_search}$ is the clause generation function following Algorithm 1, 
$f_\mathit{enumerate}$ is the fact enumeration function following Algorithm 2, 
$\mathcal{W}$ is the set of weights, 
and $T$ is the time step for the infer function.

We solve ILP problem $\mathcal{Q}$ by minimizing cross-entropy loss, defined as:
\begin{align}
    \mathit{loss} = -\mathbb{E}_{(E, y) \sim \mathcal{Y}} [ &y \log  p(y ~|~ E, \mathcal{Q}, \mathcal{C},\mathcal{W}, T) + \nonumber\\ & (1-y) \log (1 -  p(y ~|~ E, \mathcal{Q}, \mathcal{C},\mathcal{W}, T))].
\end{align}

\section{Experiments}
In this section, we experimentally support the following claims:
    (1) Our enumeration algorithm yields a reasonable number of ground atoms.
    (2) Our clause generation algorithm improves the performance of differentiable program searching.
    (3) Our soft program composition is efficient in terms of memory and computation costs.
    (4) Our framework learns logic programs successfully from noisy and structured examples, which are outside the scope of both $\partial$ILP and standard ILP approaches.

\noindent We performed our experiments\footnote{The source code of all experiments will be available at {\tt https://github.com/hkrsnd/dilp-st}} on several standard ILP tasks with structured examples, partially adopted from Shapiro and Caferra ~\cite{Shapiro83,Caferra13}.
Through all the tasks, sets of variables were consistently fixed, i.e.,  $\mathcal{V} = \{x,y,z,v,w\}$.
\\
 {\bf Member} The task is to learn the membership function for lists.  
 The language is given as $\mathcal{P}=\{ mem/2 \}$, $\mathcal{F}=\{f/2\}$, $\mathcal{A} = \{a,b,c,* \}$.
 The initial clause is $\mathcal{C}_0 = \{ \mathit{mem}(x,y) \}$.
The problem is briefly described:
 \begin{align*}
   \mathcal{E}^+ &= \{\mathit{mem}(a, [a, c]), \mathit{mem}(a, [b, a]), \ldots \},\\
    \mathcal{E}^- &= \{ \mathit{mem}(c, [b, a]), \mathit{mem}(c, [a]), \ldots \},\\
    \mathcal{B} &= \{\mathit{mem}(a, [a]), \mathit{mem}(b, [b]), \mathit{mem}(c, [c]) \}.
\end{align*}
 {\bf Plus}  The task is to learn the plus operation for natural numbers.
   The language is given as $\mathcal{P}=\{ plus/3 \}$, $\mathcal{F}=\{s/1\}$, $\mathcal{A} = \{0 \}$.
   The initial clause is $\mathcal{C}_0 = \{ \mathit{plus}(x,y,z) \}$.
The problem is briefly described:
\begin{align*}
    \mathcal{E}^+ &= \{\mathit{plus}(s(0), 0, s(0)),  \mathit{plus}(s^5(0), s^3(0), s^8(0)), \ldots \},\\
    \mathcal{E}^- &= \{\mathit{plus}(s(0), s^2(0), 0), \mathit{plus}(0, s^2(0), s^4(0)), \ldots \},\\
    \mathcal{B} &= \{ \mathit{plus}(0,0,0) \}.
\end{align*}
 {\bf Append} The task is to learn the append function for lists. 
    The language is given as $\mathcal{P}=\{ app/3 \}$, $\mathcal{F}=\{f/2\}$, $\mathcal{A} = \{a,b,c,*  \}$. 
    The initial clause is $\mathcal{C}_0 = \{ \mathit{app}(x,y,z) \}$.
The problem is briefly described:
    \begin{align*}
    \mathcal{E}^+ &= \{\mathit{app}([c], [], [c]), \mathit{app}([a,a,b], [b,c], [a,a,b,b,c]), \ldots \},\\
    \mathcal{E}^- &= \{ \mathit{app}([], [a,a], [a,a,b]), \mathit{app}([b], [], [c]), \ldots \},\\
    \mathcal{B} &= \{\mathit{app}([],[],[]) \}. 
\end{align*}
{\bf Delete}  The task is to learn the delete operation for lists. 
The language is given as $\mathcal{P}=\{ del/3 \}$, $\mathcal{F}=\{f/2\}$, $\mathcal{A} = \{a,b,c,* \}$.
The initial clause is $\mathcal{C}_0 = \{ \mathit{del}(x,y,z) \}$.
The problem is briefly described:
\begin{align*}
    \mathcal{E}^+ &= \{\mathit{del}(b, [a, c, b], [a, c]), \mathit{del}(a, [b, a, a], [b, a]), \ldots \},\\
    \mathcal{E}^- &= \{ \mathit{del}(c, [c, a, a], [a, b]), \mathit{del}(b, [b], [a]), \ldots \},\\
    \mathcal{B} &= \{ \mathit{del}(a, [a], []), \mathit{del}(b, [b], []), \mathit{del}(c, [c], []) \}.
\end{align*}
 {\bf Subtree} The task is to learn the subsumption relation for binary trees. 
 The language is given as $\mathcal{P}=\{ sub/2 \}$, $\mathcal{F}=\{f/2\}$, $\mathcal{A} = \{a,b,c \}$.
 The initial clause is $\mathcal{C}_0 = \{ \mathit{sub}(x,y) \}$.
 The problem is briefly described:
  \begin{align*}
    \mathcal{E}^+ &= \{\mathit{sub}(f(b,b),f(f(f(b,b),f(a,c)),f(a,c))),  \ldots \},\\
    \mathcal{E}^- &= \{ \mathit{sub}(f(a,a),f(f(c,a),f(a,c))), \ldots \},\\
    \mathcal{B} &= \{\mathit{sub}(a,a), \mathit{sub}(b,b), \mathit{sub}(c,c) \}.
\end{align*}  
In each task, we randomly generate $50$ examples for each class.
Note that the list objects are represented in a readable form, e.g., term $f(a, f(b, *))$ is represented as $[a,b]$.

\subsection{Experimental Methods and Results}
\paragraph{Hyperparameters}
To generate clauses, we used several biases for them: (i) the maximum number of bodies, denoted by $N_\mathit{body}$, and (ii) the maximum number of the nests of function symbols, denoted by $N_\mathit{nest}$. 
In all experiments, we set $N_\mathit{body} = 1$ and $N_\mathit{nest} = 1$.
We set beam size $N_\mathit{beam}$, and beam step $T_\mathit{beam}$ is  
$(N_\mathit{beam}, T_\mathit{beam}) = (3,3)$ for the Member task, $(N_\mathit{beam}, T_\mathit{beam}) = (15,3)$ for the Subtree task, and $(N_\mathit{beam}, T_\mathit{beam}) = (10,5)$ for the other tasks.

We set target program size $m$ as $m=2$ for the Member and Delete tasks, $m=3$ for the Plus and Append tasks, and $m=4$ for the Subtree task.
We set $T$ for the differentiable inference as $T=8$ for the Plus task and $T=4$ for the other tasks.
We set $\gamma = 10^{-5}$ for the softor function.

We trained our model with the RMSProp optimizer with a learning rate of $0.01$ for $3000$ epochs. 
We sampled mini-batches during the optimization, and each mini-batch contained $5\%$ of the training examples chosen randomly for each iteration.
The weights were initialized randomly in each trial.
We divided the data into $70\%$ training and $30\%$ test.
All experiments were performed on a desktop computer using its GPU\footnote{CPU: Intel(R) Xeon(R) CPU E5-1650 v4 @ 3.60 GHz, GPU: GeForce 1080Ti 11 GB, RAM: 64 GB}.

\begin{table}[t]
\small
    \centering
    \begin{tabular}{ccccc}
         Member & Plus & Append & Delete & Subtree \\\hline
         $228$ &  $1857$ & $2899$ &  $2513$ & $2172$ 
    \end{tabular}
    \caption{Number of enumerated ground atoms}
    \label{table:ground_atoms}
\end{table}

\noindent {\bf Experiment 1}
To support claim 1, we show the number of enumerated ground atoms for training data in each dataset in Table \ref{table:ground_atoms}.
Our enumeration algorithm yielded a reasonable number of ground atoms in each dataset.
The $\partial$ILP approach is infeasible in our setting because, although it considers all the ground atoms generated in the language, an infinite number of them can be generated with function symbols.

\noindent {\bf Experiment 2}
To support claim 2, we compared $two$ clause generation algorithms:
(i) generation by beam searching and refinement and (ii) naive generation without beam searching.
In setting (ii), we generated clauses without evaluation by examples. Like $\partial$ILP, it did not use the given examples during clause generation.
We set a number of clauses, denoted by $N_\mathit{clause}$. The generation stopped when the number of generated clauses exceeded $N_\mathit{clause}$.
We performed classification with different $N_\mathit{clause}$. We changed the value from $10$ to $40$ by increments of $10$ and ran the experiments $5$-times with random-weight initialization.

\begin{figure}[t]
  \begin{center}
    \begin{tabular}{c}
      \begin{minipage}{0.5\hsize}
        \begin{center}
          \includegraphics[clip, width=\linewidth]{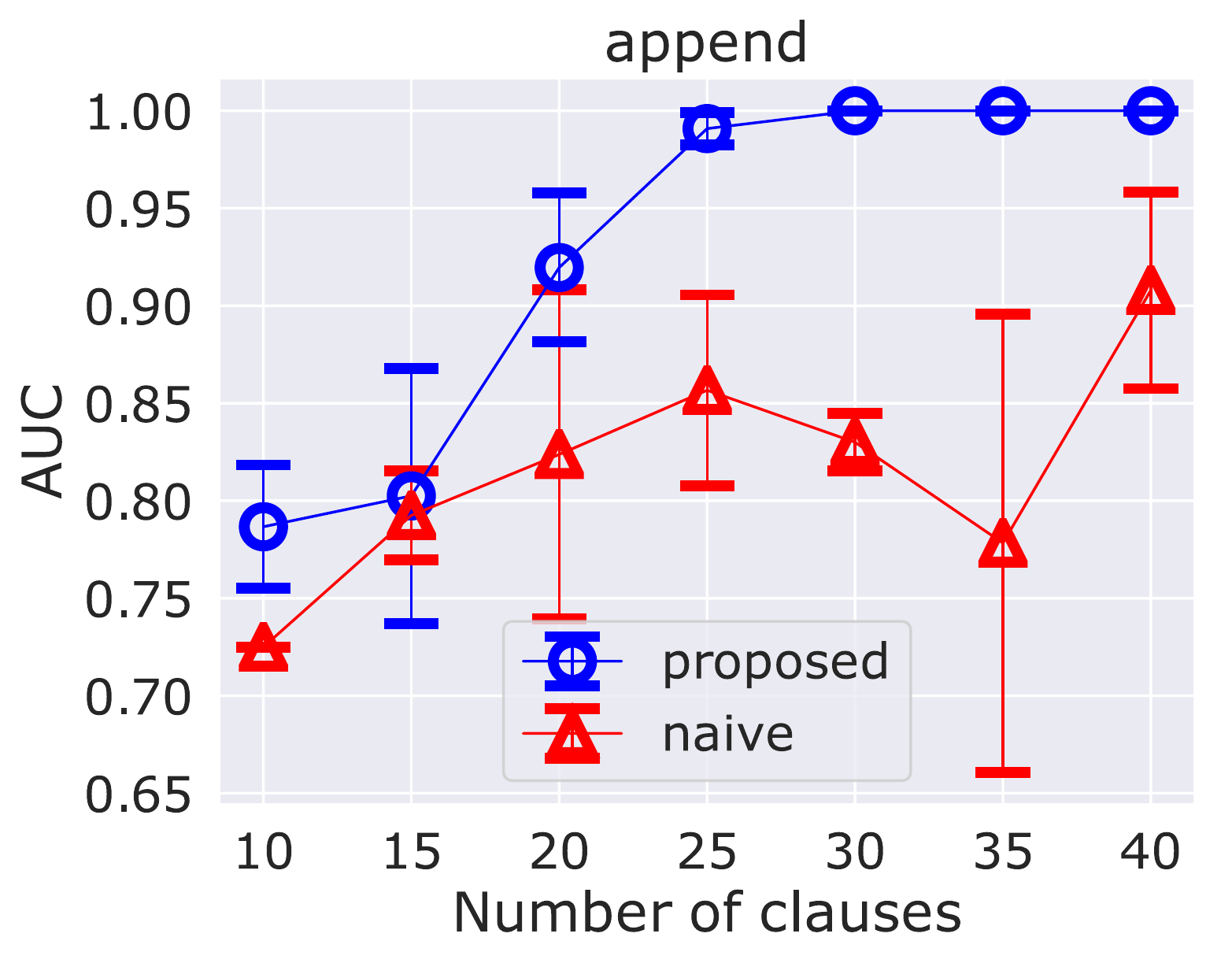}
        \end{center}
      \end{minipage}
        \begin{minipage}{0.5\hsize}
        \begin{center}
          \includegraphics[clip, width=\linewidth]{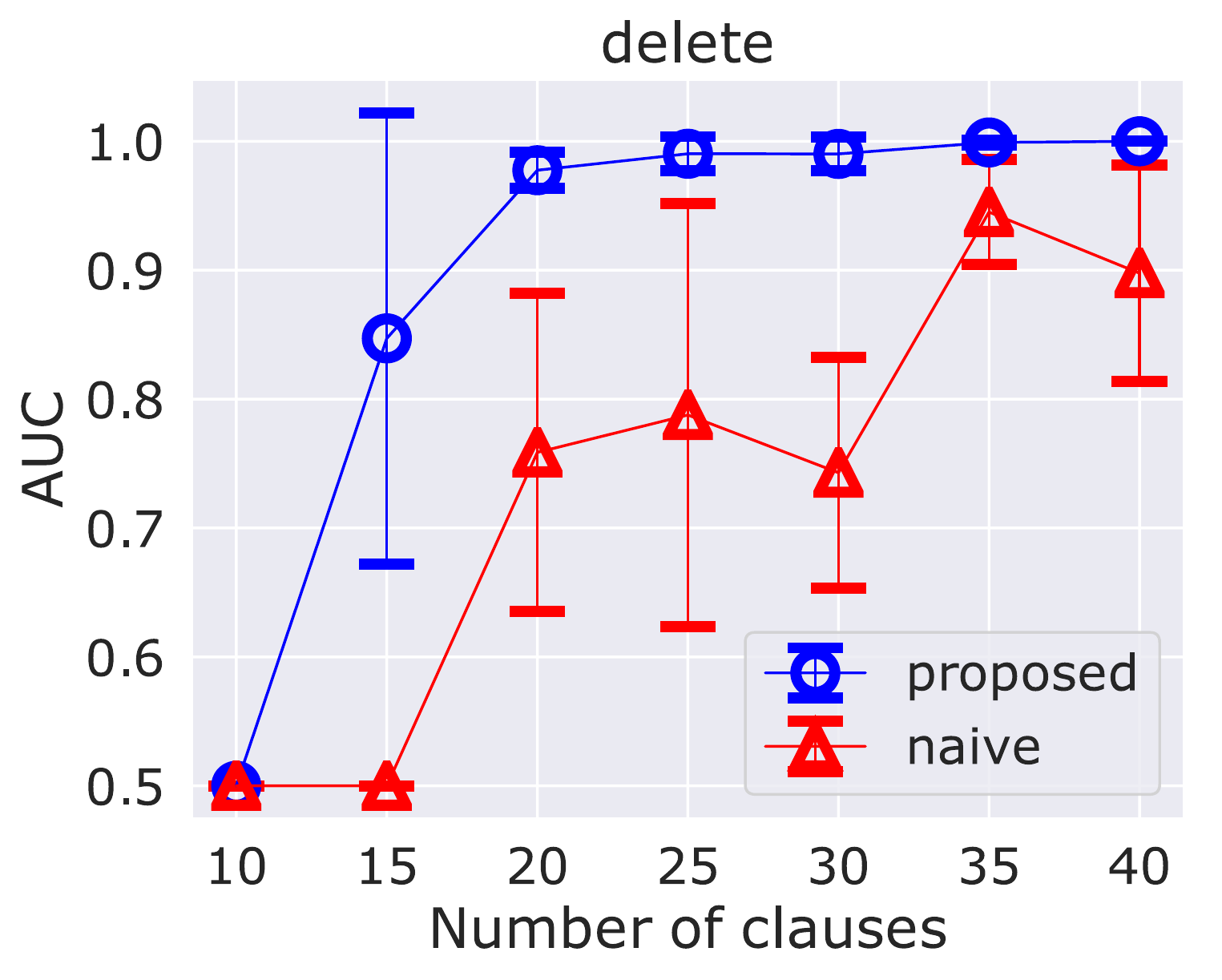}
        \end{center}
      \end{minipage}
    \end{tabular}
    \end{center}
  \caption{AUC for number of generated clauses}
  \label{fig:generation}
\end{figure}

Figure \ref{fig:generation} shows the AUC for the Append and Delete tasks. 
In each task, our approach achieved AUC scores of $1.0$ with fewer clauses.
These results show that our clause generation algorithm improved the differentiable solver, i.e., yielded an efficient search space.

\noindent {\bf Experiment 3}
To support claim 3, we compared $2$ different approaches for the infer function:
(i) multiple weights and softor approach (proposed here) and (ii) $2$-d weights for pairs of clauses.
Setting (ii) is a $\partial$ILP approach, which defines a probability distribution over the pairs of clauses $\mathcal{C}$, i.e., we assigned weights in the form of a $2$-d matrix ${\bf W} \in \mathbb{R}^{|\mathcal{C}| \times |\mathcal{C}|}$.
We compared the number of parameters and mean runtimes for each step of the gradient descent.

Table \ref{tab:expr3} shows our results. In each dataset, the proposed approach had fewer parameters.
Moreover, the mean runtime of the gradient descent was much shorter than with the pairing approach.
These results show that our approach was efficient in terms of memory and computation costs.

\begin{table}[t]
\small
        \begin{center}
            \begin{tabular}{ccccc}
            & \multicolumn{2}{c}{Parameters} & \multicolumn{2}{c}{Runtime [s]}\\
            & Proposed & Pair & Proposed & Pair \\  \hline
            Member & $24$ & $144$  & ${\bf 0.015}$ & $0.12$ \\ 
            Plus &  $120$ & $1600$   & ${\bf 0.03 }$ & $6.91$ \\ 
            Append & $150$ & $2500$  & ${\bf 0.09 }$ & $5.18$ \\
            Delete &  $150$ & $2500$ &   ${\bf 0.06}$ & $5.4$ \\
            Subtree & $80$ & $400$ &  ${\bf 0.039}$ & $1.11$ \\ 
            \hline 
        \end{tabular}
    \end{center}
    \caption{Number of parameters and mean runtime in learning steps}
    \label{tab:expr3}
\end{table}

\noindent {\bf Experiment 4}
To support claim 4, we evaluated our approach by changing the proportion of the mislabeled training data.
First, we generated training examples. Then we flipped the label of examples to make noise according to the proportion.
We changed the proportion of mislabeled data from $0.0$ to $0.5$ by increments of $0.05$. 
We ran the experiments $5$-times with random-weight initialization.

\begin{figure}[t]
\begin{center}
      \begin{tabular}{c}
      \begin{minipage}{0.5\hsize}
        \begin{center}
          \includegraphics[clip, width=\linewidth]{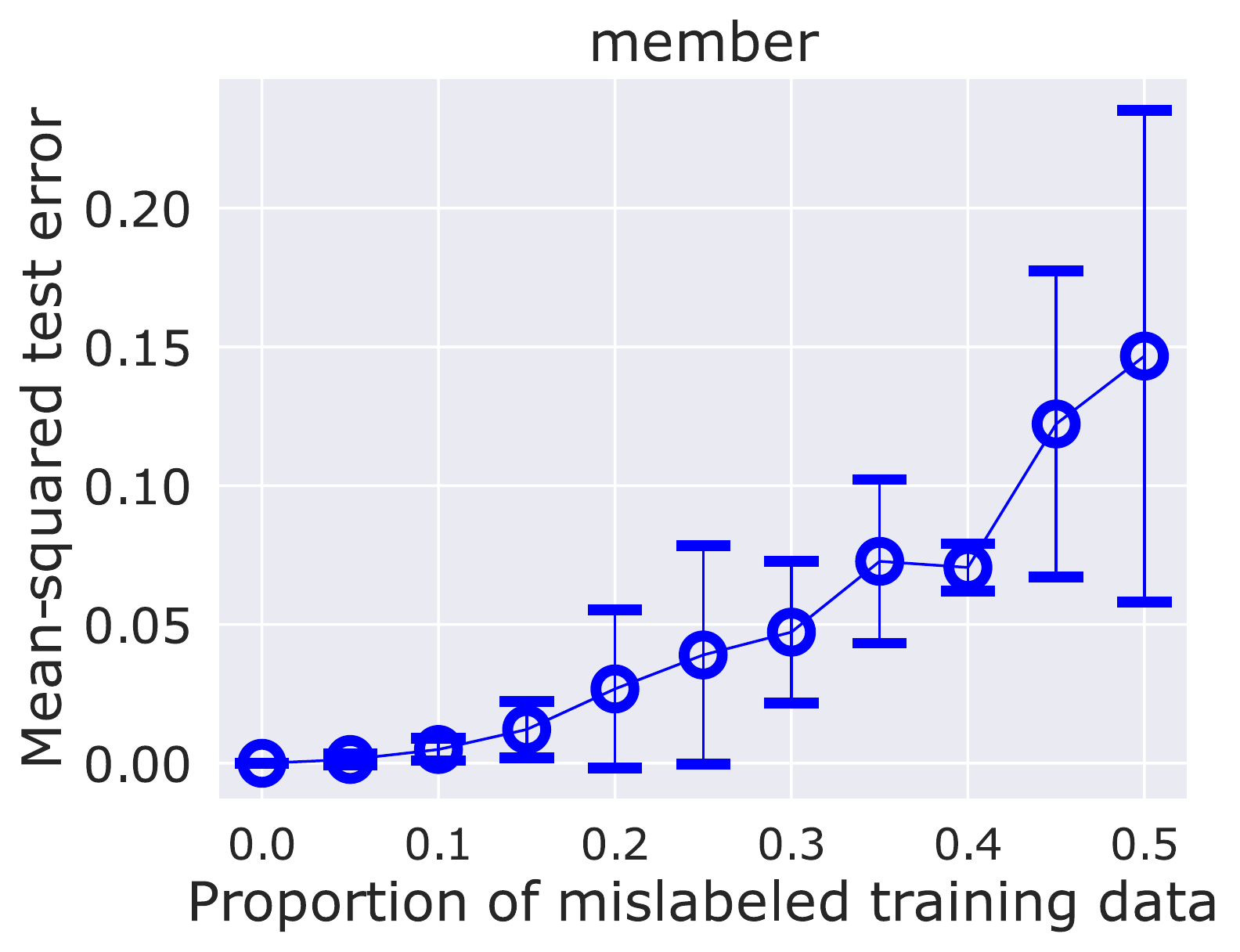}
        \end{center}
      \end{minipage}
      \begin{minipage}{0.5\hsize}
        \begin{center}
          \includegraphics[clip, width=\linewidth]{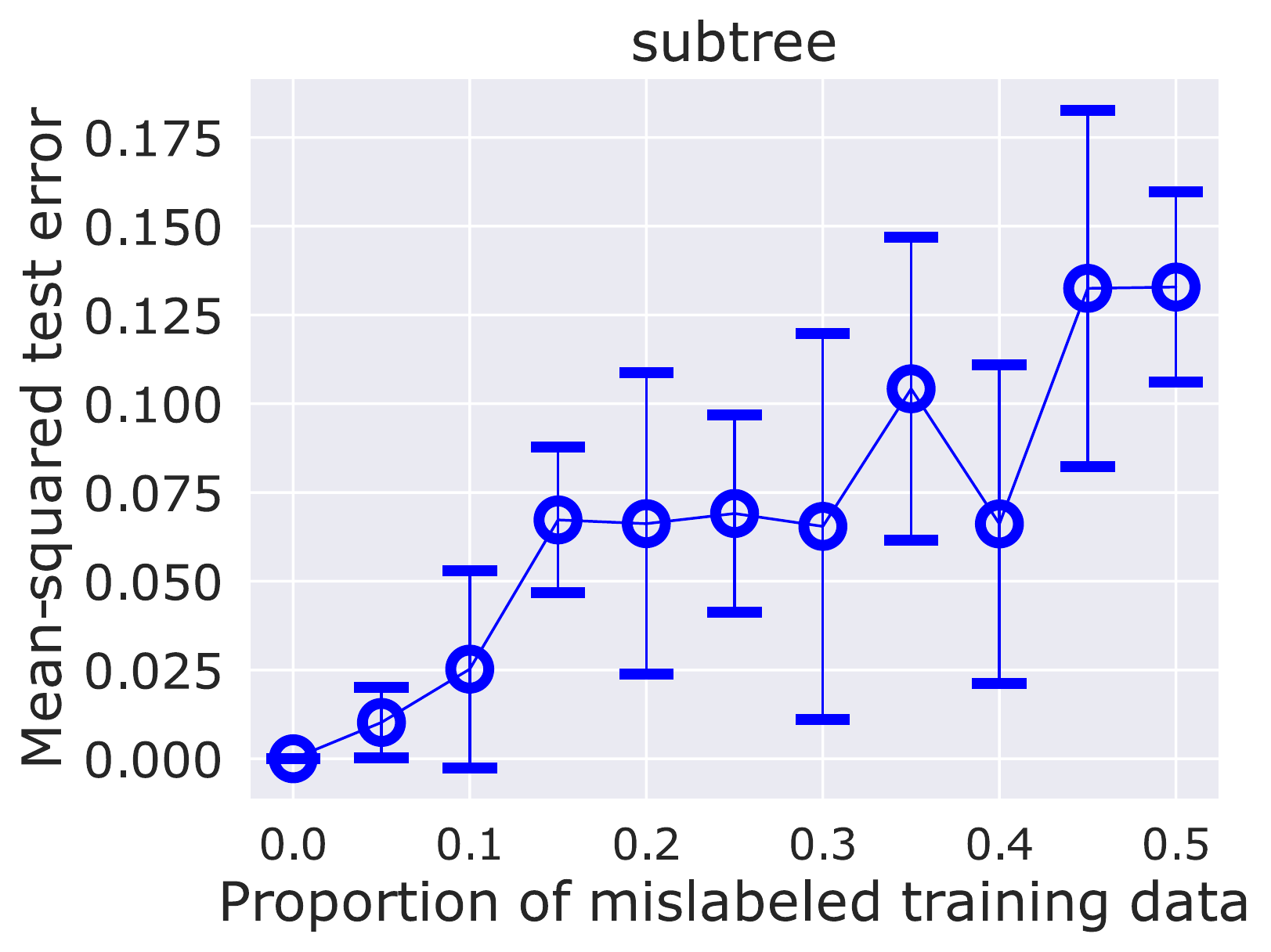}
        \end{center}
      \end{minipage}
    \end{tabular}
    \end{center}
    \caption{Mean-squared test error as proportion of mislabeled training data}
    \label{fig:noise}
\end{figure}

Figure \ref{fig:noise} shows the mean-squared test error for the proportion of mislabeled training data in the Member and Subtree tasks.
In each task, the test error increased gradually as the noise proportion increased.
Moreover, our method achieved test error less than $0.05$ with $10\%$ mislabeled training data in both tasks.
This shows that our approach was robust to noise, i.e., it found a functional theory even if there were mislabeled data.
Note that standard ILP approaches fail to find a theory when there are mislabeled data.

We show an example of the obtained programs in Table \ref{table:expr1}. 
The clauses for lists are represented in a readable form, e.g., term $f(x,y)$ is represented as $[x|y]$.
In the Plus task, the last clause represents the plus operation considering commutativity for natural numbers.
Also the last clause in the Append task can be interpreted clearly. If $v$ is obtained by appending $y$ to $z$, then the result of appending $y$ with head $x$ to $z$ is obtained just by $v$ with head $x$.
Our framework learned structured knowledge from structured examples beyond relational logic.

\begin{table}[t]
\begin{center}

    \begin{tabular}{cc}
    Problem & Learned logic program\\ \hline\hline
        Member & \begin{tabular}{c}  $\mathit{mem}(x,[y|z]) \leftarrow \mathit{mem}(x,z)$ \\ $\mathit{mem}(x,[x|y])$ \end{tabular}\\ \hline
        Plus & \begin{tabular}{c}  $\mathit{plus}(0,x,x)$ \\ $\mathit{plus}(x,s(y),s(z)) \leftarrow \mathit{plus}(x,y,z)$\\ $\mathit{plus}(s(x),y,s(z))\leftarrow\mathit{plus}(y,x,z)$ \end{tabular}\\ \hline
        Append & \begin{tabular}{c} $\mathit{app}([], x, x)$\\ $\mathit{app}(x, [], x)$\\ $\mathit{app}([x|y], z, [x|v]) \leftarrow \mathit{app}(y,z,v) $\end{tabular}\\ \hline
        Delete & \begin{tabular}{c}  $\mathit{del}(x,[x|y],y)$ \\$\mathit{del}(x,[y|z],[y|v]) \leftarrow \mathit{del}(x,z,v)$ \end{tabular}\\ \hline
        Subtree & \begin{tabular}{c}  $\mathit{sub}(f(x,y),f(x,y))$ \\ $\mathit{sub}(x,f(y,z)) \leftarrow sub(x,z)$\\ $\mathit{sub}(x,f(y,z)) \leftarrow sub(x,y)$ \\ $\mathit{sub}(x, f(y,x))$  \end{tabular} \\\hline
    \end{tabular}
    \caption{Learned logic programs in standard ILP tasks}
    \label{table:expr1}
\end{center}
\end{table}

\section{Conclusion}
We proposed a new differentiable inductive logic programming framework that
deals with complex logic programs with function symbols that yield readable outputs for structured data.
To establish our framework, we proposed $three$ main contributions.
First, we proposed a clause generation algorithm that uses beam searching with refinement. 
Second, we proposed an enumeration algorithm for ground atoms. 
Third, we proposed a soft program composition approach using multiple weights and the softor function. 

In our experiments, we showed: 
(i) our enumeration algorithm yields a reasonable number of ground atoms,
(ii) our clause generation algorithm improves the performance of differentiable program searching, 
(iii) our soft program composition is efficient in memory and computation costs, and 
(iv) our framework learns logic programs successfully from noisy and structured examples, which are outside the scope of both $\partial$ILP and standard ILP approaches.

One major limitation of our framework is its scalability for large-scale programs.
A high-quality search space is necessary to deal with more expressive programs, such as sorting.
Further research could tackle this problem by incorporating such declarative bias~\cite{Claire96,DeRaedt12} as mode declarations~\cite{Muggleton91} and metarules~\cite{Cropper16} to manage the search space.

To the best of our knowledge, this is the first work that incorporates symbolic methods, such as refinement, with a differentiable ILP approach.
We believe that our work will trigger future work to combine the best of both the symbolic and subsymbolic worlds.

\subsubsection*{Acknowledgments}
This work was partly supported by JSPS KAKENHI Grant Number
17K19973.

\bibliography{main}

\end{document}